# Leveraging Human Feedback to Scale Educational Datasets Combining Crowdworkers and Comparative Judgement

**Owen Henkel**
University of Oxford
Oxford, United Kingdom
owen.henkel@gmail.com

**Libby Hills**
Jacobs Foundation
Zurich, Switzerland
olivia.lh.hills@gmail

## ABSTRACT

Machine Learning models have many potentially beneficial applications in education settings, but a key barrier to their development is securing enough data to train these models. Labelling educational data has traditionally relied on highly skilled raters using complex, multi-class rubrics, making the process expensive and difficult to scale. An alternative, more scalable approach could be to use non-expert crowdworkers to evaluate student work, however, maintaining sufficiently high levels of accuracy and inter-rater reliability when using non-expert workers is challenging. This paper reports on two experiments investigating using non- expert crowdworkers and comparative judgement to evaluate complex student data. Crowdworkers were hired to evaluate student responses to open-ended reading comprehension questions. Crowdworkers were randomly assigned to one of two conditions: the control, where they were asked to decide whether answers were correct or incorrect (i.e., a categorical judgement), or the treatment, where they were shown the same question and answers, but were instead asked to decide which of two candidate answers was more correct (i.e., a comparative/preference-based judgement). We found that using comparative judgement substantially improved inter-rater reliability on both tasks. These results are in-line with well-established literature on the benefits of comparative judgement in the field of educational assessment, as well as with recent trends in artificial intelligence research, where comparative judgement is becoming the preferred method for providing human feedback on model outputs when working with non-expert crowdworkers. However, to our knowledge, these results are novel and important in demonstrating the beneficial effects of using the combination of comparative judgement and crowdworkers to evaluate educational data.

## 1. INTRODUCTION

Using artificial intelligence (AI) machine learning (ML) models in education can have many potential benefits, including inferring student knowledge, computer adaptive testing, predicting possible school dropouts and providing insight into drivers of student boredom [1], to name just a few. A key hurdle to developing models is having enough labelled data (i.e., graded student responses) necessary to fine-tune models, which typically require thousands of examples (at a minimum) to reach acceptable levels of performance. Creating large, labelled datasets of educational data is challenging as most education data requires complex, holistic or subjective judgements to be made (e.g., evaluating student responses to open-ended questions) [2]. For these types of tasks, raters must use judgement, making inter-rater reliability (IRR) the key measure of data quality [3]. Achieving sufficiently high IRR is traditionally achieved by using expert raters to assign rubric-based scores, which can be expensive and time-consuming.

A potentially promising approach for labelling large-scale data sets for educational AI models is working with crowdworkers but asking them to make preference-based or comparative judgements, rather than categorical ratings. Research from machine learning and the social sciences suggests that this approach can be easier for raters and improve inter-rater reliability [4,5]. Reinforcement learning from human feedback (RLHF) led to a decisive improvement in ChatGPT and is now rapidly being adopted as an approach [6]. RLHF is based on the insight that human feedback is critical to improving model outputs when aspects of these outputs are hard to define (for example, how helpful, honest, or harmful outputs are [6]. Researchers have begun to converge on the benefits of asking human raters to make relative judgements about which output they preferred (called preference-based rating). Making relative judgements is easier for raters when hard-to-define criteria are being used to rate outputs and generally leads to high inter-rater reliability [7].

Interestingly, this dovetails with well-established work from educational psychology, which has established that there can be numerous benefits to making relative judgements when evaluating student work. Comparative judgement (the term used for making relative rather than absolute judgements) has a long history in psychology and is widely understood to improve rater accuracy and inter-rater reliability, lower training requirements and be less cognitively strenuous for raters than making absolute judgements [8]. Comparative judgement involves asking raters to compare and rank two or more pieces of work, rather than assigning them a score based on a predetermined scale or standard (as raters need to do when making absolute judgements) [7]. While there is substantial research both on using comparative judgement to evaluate educational data, and in using preference-based rating approaches when asking crowdworkers to make complex and or holistic judgements, there is very little research that combines these approaches. This paper aims to address that gap.

## 2. PRIOR WORK

**Crowdsourcing**, which involves "the outsourcing of a piece of work to a crowd of people via an open call for contributions" is commonly used in machine learning research [3]. The benefits of crowdsourcing are numerous, including speed, scale, the ability to obtain judgments from much larger and more representative portions of the population, and the potential to harness the 'wisdom of crowd' effect [9].Crowdsourcing can include both paid and volunteer work and be done by experts or non-experts, but as crowdsourcing has become a more established method, the practice of contracting individuals to complete labelling, evaluation, and other tasks on online platforms, also known as crowdwork, has become increasingly common [4]. However, labelling examples of more complex tasks, such as evaluating the truthfulness or toxicity of a statement, or determining which product review is more helpful, are both more time-consuming to complete, and more difficult to achieve high levels of accuracy and reliability [10]. This has raised questions about how to improve the accuracy and inter-rater reliability of non-expert evaluation of these more complex tasks. Various ways to improve the accuracy of human judgements on complex tasks have been proposed, including offering raters initial practice and feedback, financial incentivization, pooling of judgments, and selectively screening raters to ensure attentiveness and quality [11, 12]. Asking raters to make relative rather than absolute judgements (preference-based rating in machine learning or comparative judgement in educational psychology) is another way to improve rater accuracy and inter-rater reliability.

Recent research has used reinforcement learning from **human feedback** (RLHF) to improve model performance across a range of tasks [13, 6]. In the case of InstructGPT, one of the models behind chat GPT, RLHF was used during fine-tuning to better align model output with the type of response that users were likely to prefer. As part of this approach, labelers and researchers were asked to rank model outputs from best to worst based on which response they preferred. This data was then used to train a reward model, which was used to fine-tune the model to the stated preferences of the human labelers [6]. Incorporating crowdworkers preferences (i.e. asking them to make relative judgements) of the quality and appropriateness of model outputs was a key component of the increased relevance and naturalness seen in ChatGPT as compared to GPT-3, which uses the same underlying statistical model [13]. Furthermore, Stiennon et al (2022) found that comparative or preference-based approaches can lead to agreement rates between expert and non-expert raters that are almost as high as expert-to-expert agreement rates [13].

Unlike absolute or categorical judgments, **comparative judgement** asks evaluators to make relative comparisons between two or more pieces of work, rather than scoring them against a predetermined standard. Psychology research dating back to Thurstone's law of comparative judgement has consistently shown that humans are better at making comparative ratings than they are at scale or absolute ratings (for example estimates of quantity, the intensity of sound or the order of weights [14, 15, 16]. This finding has now been echoed by recent machine learning research on reinforcement learning from human feedback [13]. In educational assessment, evidence suggests that using multiple pairwise ratings can result in very high reliability, equal to or exceeding area experts who undergo extensive calibration exercises [7]. This is especially the case for complex judgements that require a degree of holistic judgement, like evaluating open-response answers to reading comprehension questions [7]. Other research has also shown that the accuracy of expert raters of examination scripts can be higher when making comparative rather than absolute judgements [17]. Steedle & Ferrara (2016) point to three additional advantages of comparative judgement: improved accuracy, lower training requirements and higher efficiency [8].

## 3. CURRENT STUDY
### 3.1 Overview
We conducted two experiments intended to establish proof-of-principle for using crowdworkers to label educational data. In these experiments crowdworkers were hired to evaluate student responses to two differ methods of assessing reading ability. Our goal was to investigate both the overall accuracy and agreement levels of non-expert raters, and to test whether asking them to evaluate student responses comparatively rather than categorically, improved accuracy and inter-rater reliability.

To answer this question raters were assigned to one of two conditions: categorical judgement or comparative judgement. In the categorical judgement condition, the raters were asked to make an absolute judgement about a piece of student work. In the comparative condition, a different group of raters were given the same task and were asked to decide which answer was more correct. In both conditions, the tasks were identical, with the only difference being the type of judgement they were asked to make. Across both experiments, a total of approximately 300 crowdworkers were recruited through the online research platform Prolific. This experimental design allowed us to control for any unintended differences in difficulty between correct and incorrect candidate answers, as well as the impact of the order of questions on the cognitive load associated with the task. The assignments were also quasi-random, and the tasks were posted simultaneously on the Prolific platform, making it likely that the difference in accuracy and reliability levels is attributable to the differences between comparative and categorical judgments.

### 3.2 Rating Tasks and Datasets
We used asked rater to evaluate student responses to two commonly used methods for assessing reading ability (a) short-answer responses to reading comprehension questions, and (b) oral reading fluency. These specific tasks were selected both because they are widely used for formative assessment of reading ability and because they are time-consuming for teachers to grade. We also selected these two tasks because they differ in the complexity of the rating process. While evaluating short-answer responses reading comprehension questions is a relatively straightforward task, evaluating oral reading fluency, in particularly prosody, is considered a task that required highly trained educators.

For the short-answer task, approximately 40 raters were recruited. They evaluated 10 examples, each one consisting of a short nonfiction passage, a reading comprehension question, and a candidate answer. In the categorical judgement condition, the raters were asked to decide if the candidate answer was correct or incorrect (equivalent to a two-way classification). In the comparative condition, a different group of raters were given the same passage, a question, and two candidate answers from the curated set (one correct and one incorrect) and were asked to decide which answer was more correct. In both conditions the passages, questions and candidate answers were identical and the only difference was the type of judgement they were asked to make. The passages and questions for this task were taken from SQUAD 2.0, a corpus of passages taken from Wikipedia, each accompanied by a direct question about the passage, and a correct answer.

| |
|---|
| **Passage**: Soon after the Normans began to enter Italy, they entered the Byzantine Empire and then Armenia, fighting against the Pechenegs, the Bulgars, and especially the Seljuk Turks. Norman mercenaries were first encouraged to come to the south by the Lombard's to act against the Byzantines, but they soon fought in Byzantine service in Sicily, and then in Greek service. They were prominent alongside Varangian and Lombard contingents in the Sicilian campaign of George Maniacs in 1038 AD. |
| **Question:** Who was the Normans' main enemy in Italy, the Byzantine Empire and Armenia? |

| Experimental Conditions ||
|---|---|
| **Categorical** | **Comparative** |
| Based on the passage is this a correct answer to the question?<br><br>-------------------------------<br><br>"The Seljuk Turks, as well as the Pechenegs" **(a) yes**     **(b) no**<br><br>----------------------------- | Based on the passage which answer is more correct?<br><br>-------------------------------<br><br>**(a) The Seljuk Turks, as well as the Pechenegs** |
| The Greeks, Italian, and Sicilians.<br>(a) yes  **(b) no** | (b) The Greeks, Italian, and Sicilians. |

**Fig. 1. Example of short-answer rating task**

For the oral reading fluency task, approximately 250 raters were recruited through the Prolific platform and asked to evaluate 44 audio clips of students reading short stories for prosody, or the overall expressivity and naturalness of their reading. The audio files were used with permission from CU Kid's Read and Summarized Story Corpus. In the comparative condition, raters were given two examples of students reading, along with a detailed description of prosody, and were asked to decide which of the two audio clips had better prosody. In categorial conditions, raters were given individual audio clips of students reading, along with a well-validated rubric, and were asked to assign a score (i.e., classify) to the audio clip into one of two categories.

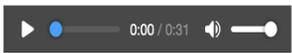

**Fig. 2 Example of prosody rating task**

### 3.3 Rater Selection and Quality Control

As prior research indicated that the rater's native language and previous task completion success rate were associated with performance on the rating tasks, we recruited candidates from the US, UK, and Canada who had a historically successful completion rate of at least 85%. We also employed best practice quality control measures, including embedded attention checks and minimum completion time to ensure raters were diligent [4, 5]. While the use of crowdworkers can be an efficient and cost-effective method for data annotation, prior research highlighted the importance of researchers taking steps to ensure that crowdworkers are treated fairly [18]. For this study raters were recruited from the online platform Prolific, known for its ethical practices including dispute resolution protocol, and were paid a benchmarked rate of GBP 8.5 per hour.

### 3.4 Reliability Measures

A critical consideration for labelling and using educational data is ensuring that a given approach has high reliability, i.e., that the data would receive the same rating across various settings, timeframes, and by distinct raters. The simplest method for evaluating reliability with nominal data involves calculating the observed agreement. However, this metric does not account for the anticipated agreement based on chance, causing a bias towards evolution tasks with fewer categories [19]. To circumvent this issue, alternative reliability measures have been introduced, most notably Cohen's kappa [20], which corrects observed agreement for chance agreement. However, Cohen's kappa is restricted to the unique case of two raters, an uncommon scenario in large-scale data annotation. Consequently, another measure, Krippendorff's alpha [21] was developed, offering significant flexibility regarding the measurement scale and the number of raters. This has become the favored metric for assessing inter-rater reliability in various data labelling tasks. The value of Krippendorf's alpha can range from -1 to +1 and can be interpreted similarly to a correlation coefficient, with -1 being perfect disagreement, 0 being complete random chance, and 1 being perfect agreement. Adequate reliability benchmarks vary but are typically above 0.5. Landis and Koch [21] define substantial as reliability coefficients larger than 0.6 as substantial, and Krippendorff claims that alpha scores between 0.67 and 0.80 can be used for drawing provisional conclusions.

### 3.5 Significance Testing

In the context of this study there are two questions relating to statistical significance: (a) are the reliability levels themselves statistically significant, and (b) is any observed difference in the reliability levels based on the methods of rating (i.e., categorical and comparative) significant?

For reliability measures, the confidence interval establishes a range within which the actual coefficient is likely to be found with a specified probability and hence can be used for hypothesis testing. For instance, if the goal is to demonstrate reliability score is greater than chance at a 95% confidence level, the lower bound of the two-sided 95% confidence interval must exceed 0. Because Krippendorf's alpha already accounts for the level of agreement due to chance alone, in nearly all situations where even a minimally small sample is used and the alpha value is positive, the results are statistically significant [23].

Perhaps more interestingly, if the objective is to demonstrate that the reliability exceeds a specific benchmark then the lower limit of the confidence interval must be greater than this benchmark. For instance, to demonstrate significant reliability (as defined by Landis and Koch), the lower limit of the confidence interval would need to be greater than alpha = 0.6. Confidence intervals can also be used for hypothesis testing to determine if the difference between two agreement levels is statistically significant. If the two confidence intervals do not overlap then the null hypothesis is rejected.

For Krippendorff's alpha, the theoretical distribution of values necessary to calculate the confidence interval is not known [23] However, the empirical distribution can be obtained by the bootstrap approach, which Krippendorff proposed an algorithm for in the original paper. This method has been used by various other researchers, and we adopt the implementation and parameters laid out by Zapf et al [23].

## 4. RESULTS
### 4.1 Short Answer
To evaluate the accuracy of the ratings we compared the crowdworker answer with the correct answer to the question as determined by ourselves. When using categorical judgment, the crowdwokers had an accuracy level of was 73%, which improved to 86% when they were presented the same student responses but were asked to make comparative judgment.

The results for the reliability of short-answer ratings are summarized in Table1. Krippendorf's alpha is interpreted similarly to most other inter- rater reliability scores, with 0 being perfect disagreement, and 1 being perfect agreement, with scores above 0.6 typically overlap. considered to have substantial reliability. Raters asked to rate responses categorically achieved an alpha of .66, whereas raters asked to rate responses comparatively had an alpha of .80. confidence interval equivalent to $p < .01$; a more stringent benchmark than the conventional $p < .05$. The 0.14 increase in. We can have high confidence in these reliability scores, based on the 99% inter-rater reliability is both large and highly statistically significant, as the respective 99% confidence intervals do not overlap.

**Table 1: IRR of short-answer score, by rating type**

| Rating Type | Point Estimate (Krippendorf's alpha) | 99% confidence interval |
|---|---|---|
| Categorical | 0.66 | 0.64 - 0.67 |
| Comparative | 0.80 | 0.78 - 0.82 |
| Change | + 0.14 | |

### 4.2 Oral Fluency
Because evaluating the prosody of students reading, does not have an objectively correct score, it was not possible to calculate absolute accuracy for this tasks. The results for the reliability of the rating are summarized below in Table 2. Raters asked to evaluate the prosody of students' reading achieved had an inter-rater reliability score of 0.7, whereas raters asked to rate responses comparatively had an alpha of 0.78. We can have high confidence in these IRR estimates, based on the 99% confidence interval. The 0.08 increase in the reliability score is moderately large and highly statistically significant, as the respective 99% confidence intervals do not overlap.

**Table 2: IRR of Oral Fluency score, by rating type**

| Rating Type | Point Estimate (Krippendorf's alpha) | 99% confidence interval |
|---|---|---|
| Categorical | 0.7 | 0.77 - 0.79 |
| Comparative | 0.78 | 0.68 - 0.73 |
| Change | + 0.08 | |

### 4.3 Interpretation
The improvements in accuracy and reliability reported above can be considered substantial highly statistically significant. In the case of the accuracy of ratings for the short answer tasks, the floor for accuracy would be 50% (i.e., the coworker randomly guessing). Hence the 13-percentage point improvement in accuracy resulting from shifting the task from comparative to categorical, is ¼ of the total possible improvement on this task – moving from pure chance to perfect performance. The improvements to interrater reliability, 0.14 and 0.08, respectively, are highly statistically significant add larger than the increases in IRR produces by other commonly used methods to improve the quality crowd workers ratings such as such as candidate screening, pre-training, or incentive payments [10].

In addition to the improvement being substantial, the absolute level of reliability reaches or exceed commonly used benchmarks. To put these values in context, he NAEP, a gold-standard assessment of reading ability, directly assesses prosody with highly trained expert raters using a rubric - the same rubric used by raters in this study. Various studies have reported inter-rater reliability ranging between 0.0 and 0.80 [24]. Hence the results of these two experiments suggest that (a) by structuring rating tasks as comparative rather than categorical judgments can substantially improve reliability, and (b) that crowdworkers can achieve moderate to high levels of inter-rater reliability, equivalent to those of highly trained expert raters, andb

### 4.4 Limitations
Despite the high levels of statistical significance reported above, to attribute the increase in agreement to the change of the rating task from categorical to comparative, we must assume that the items being rated were the same (which is true) and that the raters were substantially similar. While we believe that quasi-random assignment of raters controlled for rater-specific differences, the lack of truly random assignment means we cannot definitively conclude this. Another limitation is that this effect was only demonstrated on two specific tasks, so it is yet to be determined whether this effect generalizes to other tasks.

## 5. DISCUSSION AND IMPLICATIONS
In industry settings, especially within machine learning, using crowdsourcing to label data is de rigueur. However, it is much less commonly used in the social sciences, particularly in education research, to evaluate student data. This is likely due to a variety of reasons, one of the most important ones being concerns about the accuracy and reliability of non-experts evaluating student work. This study indicates (1) that it is possible to achieve high levels of inter-rater reliability with non-expert crowdworkers, (2) that reliability levels can be meaningfully increased if they are asked to make comparative rather than categorical judgements, (3) these high levels of IRR can be attained at a relatively modest cost and in a rapid timeframe.

These findings are in-line with prior research from psychology on the benefits of using comparative judgement [14, 15, 16], and with current trends in machine learning where crowdsourcing is routinely used to label complex data [4]. However, this study is one of the first, to our knowledge, to directly investigate the impact of combining these two approaches to explore the potential for education research and assessment. If our results are confirmed through further research, there would be several interrelated implications. First, it could establish proof-of-principle for crowdworkers making comparative judgements to label and/or evaluate student work. This in turn would allow for the creation of larger, more nuanced, and representative educational datasets, that could have a variety of beneficial applications.

## 6. CONCLUSION
This study aims to examine the feasibility of using crowdsourcing approaches with comparative judgement to create a scalable approach for evaluating and labelling educational data. The results from our two experiments suggest that using comparative judgement instead of categorical judgement to make holistic evaluations of complex educational data could improve accuracy and inter-rater reliability. Furthermore, it suggests that non-expert crowdworkers can label and evaluate education data with high accuracy when asked to make comparative judgments. These results are novel as they are among the first, to our knowledge, to investigate the potential of using comparative judgement with non-expert crowdworkers for evaluating educational data. These early results suggest that further research exploring the effect of combining crowdsourcing and comparative judgement would be valuable. Furthermore, findings highlight the potential to combine machine learning and educational research methods and provide a framework for further interdisciplinary research in this field.